\documentclass{article}
\usepackage[letterpaper, left=1cm, right=1.25cm, top=1cm, bottom=1.25cm]{geometry}
\usepackage{hyperref}
\usepackage{breakurl}
\usepackage[dvips]{graphicx,color}
\usepackage{amsmath}
\usepackage{amsfonts}
\usepackage{multirow}
\usepackage{fancyhdr}
\newtheorem{theorem}{Theorem}[section]

\title{Fast Color Space Transformations Using Minimax Approximations}
\author{%
        M. Emre Celebi\\
        Dept.\ of Computer Science\\Louisiana State Univ., Shreveport, LA, USA\\
        \href{mailto:ecelebi@lsus.edu}{ecelebi@lsus.edu}\\
        \\
        Hassan A. Kingravi\\
        Dept.\ of Computer Science, Georgia Institute of Technology, Atlanta, GA, USA\\
        \href{mailto:kingravi@gatech.edu}{kingravi@gatech.edu}\\
        \\
        Fatih Celiker\\
        Dept.\ of Mathematics, Wayne State University, Detroit, MI, USA\\
        \href{mailto:celiker@math.wayne.edu}{celiker@math.wayne.edu}\\
       }
\begin{document}
\maketitle

\begin{abstract}
Color space transformations are frequently used in image processing, graphics, and visualization applications. In many cases, these transformations are complex nonlinear functions, which prohibits their use in time-critical applications. In this paper, we present a new approach called Minimax Approximations for Color-space Transformations (MACT). We demonstrate MACT on three commonly used color space transformations. Extensive experiments on a large and diverse image set and comparisons with well-known multidimensional lookup table interpolation methods show that MACT achieves an excellent balance among four criteria: ease of implementation, memory usage, accuracy, and computational speed.
\end{abstract}

\section{Introduction}
\label{sec_intro}

Color space transformations are commonly used in various image processing, graphics, and visualization tasks for decoupling luminance and chromaticity information, ensuring approximate perceptual uniformity, or achieving invariance to different imaging conditions such as viewing direction, illumination intensity, and highlights. Since color devices are usually provided with direct RGB signal input and output, the RGB color space is generally the source in these transformations.

In many cases, color space transformations are complex nonlinear functions, which prohibits their use in time-critical applications. In this paper, we present a new approach called Minimax Approximations for Color-space Transformations (MACT). We demonstrate MACT on transformations between the RGB space and three popular color spaces: CIELAB, HSI, and SCT. There are various reasons behind the choice of these particular spaces. First, all three spaces decouple luminance and chromaticity information, which makes them suitable for tasks including enhancement \cite{Kao05,Pitas96}, rendering \cite{Ebert02}, noise removal \cite{Jin07,Celebi07}, segmentation \cite{Comaniciu02,Perez94,Minten01}, and object recognition \cite{Corbalan01,Gevers99}. Second, an important feature of CIELAB is its approximate perceptual uniformity, an essential requirement for expressing color differences in applications such as color quantization \cite{Wu92}, mesh denoising \cite{Fleishman03}, and maximum contrast color set design \cite{Glasbey07}. Third, HSI and SCT are designed to match human intuition, which makes them useful for manual color selection \cite{Hsiao08}. Last, the first two components of HSI are shown to be invariant to viewing direction, surface orientation, illumination direction, and illumination intensity \cite{Gevers99}.

The rest of the paper is organized as follows. Section \ref{sec_background} gives the related work. Section \ref{sec_xform} presents the use of minimax approximation theory in speeding up color space transformations and the experimental results. Finally, Section \ref{sec_conc} gives the conclusions.

\section{Related Work}
\label{sec_background}

Traditionally, color space transformations in digital imaging systems have been implemented using lookup tables (LUTs) that require some form of multidimensional interpolation. The most commonly used 3D LUT interpolation methods are trilinear, prism, pyramidal, and tetrahedral \cite{Kang06}. These methods will be explained in \S \ref{sec_exp_lut}. In this section, we briefly review the interpolation methods that appear less frequently in the literature as well as some alternative approaches.

Chang et al. \cite{Chang97} developed a method called Sequential Linear Interpolation (SLI) that uses a partially separable grid structure, which allows the allocation of more grid points to the regions where the function to be interpolated is more nonlinear. This scheme often results in more accurate transformations at the expense of increased computational cost when compared to trilinear interpolation. Kacker et al. \cite{Kacker98} proposed a wavelet based method that uses a multiscale grid structure. This method is shown to achieve lower maximum error but higher average error when compared to SLI. Gupta and Gray \cite{Gupta01} presented a method called Maximum Entropy Estimation, which is a generalization of tetrahedral interpolation. This method is shown to be more accurate than tetrahedral interpolation in general while being at best only half as fast. Hemingway \cite{Hemingway02} described a method based on n-simplexes that is slightly faster than tetrahedral interpolation. However, the accuracy of the method is not discussed in the paper.

Neural networks have been applied to the color space transformation problem in a number of studies \cite{Kang92,Tominaga93,Usui00,Vrhel03}. They have the advantages of being more flexible and requiring less memory. However, they require training and parameter tuning, and are prone to overfitting \cite{Gupta01}.

\section{Approximate Color Space Transformations}
\label{sec_xform}

\subsection{Overview of Minimax Approximation Theory}
\label{sec_minimax}

Given a function $f$, we would like to approximate it by another function $g$ so that the error ($\varepsilon$) between them over a given interval is arbitrarily small. The existence of such approximations is stated by the following theorem:

\begin{theorem}\emph{(Weierstrass)}
Let $f$ be a continuous real-valued function defined on $[a,b]$, i.e.\ $f \in C[a,b]$. Then $\forall \varepsilon > 0$ there exists a polynomial $P$ such that $\| f - P \| < \varepsilon$, i.e.\ $\forall x \in [a,b], \ \left| f(x)-P(x) \right| < \varepsilon$.
\end{theorem}

This is commonly known as the minimax approximation to a function. It differs from other methods, e.g.\ least squares approximations, in that it minimizes the maximum error ($\varepsilon$) rather than the average error:
\begin{equation}
\varepsilon = \mathop {\rm max}\limits_{x \in [a,b]} \left| f(x)-P(x) \right|
\end{equation}
A similar theorem establishes the existence of a rational variant of this method \cite{Cheney00}. Let $n \geq 0$ be a natural number and let
\begin{equation}
P_n \left( [a,b] \right) = \left\{ {a_0  + a_1 x + \, \ldots \, + a_n x^n :x \in [a,b],\;\;a_i  \in \mathbb{R},\;\;i = 0, 1, \ldots, n} \right\}
\end{equation}
be the set of all polynomials of degree less than or equal to $n$. The set of irreducible rational functions, 
$R^n _m \left( {[a,b]} \right)$, is defined as:
\begin{equation}
R^n _m \left( {[a,b]} \right) = \left\{ {\frac{{p(x)}}
{{q(x)}}:\,p(x)\, \in P_n \left( {[a,b]} \right),\,\,q(x) \in P_m \left( {[a,b]} \right)} \right\}
\end{equation}
where $p$ and $q$ have no common factors. Then \cite{Cheney00}:
\begin{theorem}
For each function $f \in C[a,\,b]$, there exists at least one best rational approximation from the class $R^n _m \left( {[a,b]} \right)$.
\end{theorem}

This theorem states the existence of a rational approximation $r^* \in R^n_m \left( {[a,b]} \right)$ to a function $f \in C[a,\,b]$ that is optimal in the Chebyshev sense:
\begin{equation}
\mathop {\rm max}\limits_{x \in [a,b]} \left| f(x)-r^*(x) \right| = \mathop{\rm dist}\left( f,R^n_m \right)
\end{equation}
 
where $\mathop{\rm dist}\left( f,R^n_m \right)$ denotes the distance between $f$ and $R^n_m \left( {[a,b]} \right)$ with respect to some norm, in our case, the Chebyshev (maximum) norm. Regarding the choice between a polynomial and a rational approximant, it can be said that certain functions can be approximated more accurately by rationals than by polynomials. Jean-Michel Muller explains this phenomenon as follows ``It seems quite difficult to predict if a given function will be much better approximated by rational functions than by polynomials. It makes sense to think that functions that have a behavior that is 'highly nonpolynomial' (finite limits at $\pm \infty$, poles, infinite derivatives, $\ldots$) will be poorly approximated by polynomials.'' \cite{Muller06}.

In this study, the Remez Exchange Algorithm is used to calculate the minimax approximations. The reader is referred to \cite{Cheney00,Muller06} for more information on the theory of minimax approximations and \cite{Fraser65} for the implementation details of the Remez algorithm.

\subsection{CIELAB Color Space}
\label{sec_cielab}

\subsubsection{Color Space Description}
\label{sec_cielab_desc}

CIELAB is an approximately uniform color space standardized by the CIE (Commission Internationale de l'Eclairage) in 1976 \cite{Plataniotis00}. The L* component represents the lightness, whereas a* and b* represent the chromaticity. The transformation between the RGB and CIELAB color spaces is comprised of three steps: (1) conversion from nonlinear R'G'B' to linear RGB, (2) conversion from linear RGB to CIEXYZ, and (3) conversion from CIEXYZ to CIELAB.

In order to convert the nonlinear R'G'B' values to the linear RGB ones, inverse gamma correction (ITU-R BT.709) is performed:

\begin{equation}
\label{equ_rgb_lin}
	\begin{array}{l}
 k' \in \left\{ {\frac{{R'}}{{255}},\frac{{G'}}{{255}},\frac{{B'}}{{255}}} \right\} \\ 
 k\, \in \left\{ {\,R,\,G,\,B\,} \right\}{\rm  = }\,\left\{ \begin{array}{l}
 \frac{{k'}}{{4.5}}\quad \quad \quad \quad \quad \quad 0 \le k' < 0.081 \\ 
 \left( {\frac{{k' + 0.099}}{{1.099}}} \right)^{\frac{1}{{0.45}}} \quad 0.081 \le k' \le 1 \\ 
 \end{array} \right. \\ 
 \end{array}
\end{equation}

The conversion from linear RGB to CIEXYZ (ITU-R BT.709) is given by:

\begin{equation}
\label{equ_rgb_xyz}
\begin{array}{l}
 X = 0.412391 R +  0.357584 G +  0.180481 B \\ 
 Y = 0.212639 R +  0.715169 G +  0.072192 B \\ 
 Z = 0.019331 R +  0.119195 G +  0.950532 B \\ 
 \end{array}
\end{equation}

Finally, the conversion from CIEXYZ to CIELAB is given by:

\begin{equation}
\label{equ_xyz_lab}
\begin{gathered}
  L^* = 116f\left( {{Y \mathord{\left/
 {\vphantom {Y {Y_0 }}} \right.
 \kern-\nulldelimiterspace} {Y_0 }}} \right) - 16 \hfill \\
  a^* \, = 500\left[ {f\left( {{X \mathord{\left/
 {\vphantom {X {X_0 }}} \right.
 \kern-\nulldelimiterspace} {X_0 }}} \right) - f\left( {{Y \mathord{\left/
 {\vphantom {Y {Y_0 }}} \right.
 \kern-\nulldelimiterspace} {Y_0 }}} \right)} \right] \hfill \\
  b^* \,\, = 200\left[ {f\left( {{Y \mathord{\left/
 {\vphantom {Y {Y_0 }}} \right.
 \kern-\nulldelimiterspace} {Y_0 }}} \right) - f\left( {{Z \mathord{\left/
 {\vphantom {Z {Z_0 }}} \right.
 \kern-\nulldelimiterspace} {Z_0 }}} \right)} \right] \hfill \\
  f(t) = \left\{ \begin{gathered}
  t^{1/3} \quad \quad \quad \quad \quad \quad\,\, t > 0.008856 \hfill \\
  7.787t + {{16} \mathord{\left/
 {\vphantom {{16} {116}}} \right.
 \kern-\nulldelimiterspace} {116}}\quad \,t \leq 0.008856 \hfill \\ 
\end{gathered}  \right. \hfill \\ 
\end{gathered} 
\end{equation}

Here, $X_0$, $Y_0$, and $Z_0$ are the tristimulus values of the reference white. For the illuminant D65 these are:

\begin{equation}
\begin{array}{l}
 X_0 =  0.950456 \\ 
 Y_0 \, =  1.0 \\ 
 Z_0 \, =  1.089058 \\ 
 \end{array}
\end{equation}

The distance between two pixels $\mathbf{x}$ and $\mathbf{y}$ in the CIELAB space is given by:

\begin{equation}
\label{equ_cielab_dist}
D_{CIELAB}(\mathbf{x},\mathbf{y}) = \sqrt {\left({L_\mathbf{x}^* - L_\mathbf{y}^*} \right)^2 + \left( {a_\mathbf{x}^* - a_\mathbf{y}^* } \right)^2  + \left( {b_\mathbf{x}^* - b_\mathbf{y}^* } \right)^2 } 
\end{equation}

\subsubsection{Approximation of the Cube-Root Function}
\label{sec_cielab_cbrt}

(\ref{equ_rgb_lin}) and (\ref{equ_rgb_xyz}) can be implemented efficiently using LUTs as commonly seen in the literature \cite{Kang06}. The cube-root function (\emph{cbrt}) in (\ref{equ_xyz_lab}) is the main factor that influences the computational time of the transformation. The probabilities of calling this function can be calculated from an image that contains every possible color in the 24-bit RGB space (see \S \ref{sec_exp_err}):

\begin{equation}
\label{equ_cbrt_prob}
\begin{array}{l}
 P\left( {{X \mathord{\left/
 {\vphantom {X {X_0 }}} \right.
 \kern-\nulldelimiterspace} {X_0 }} > 0.008856} \right) = 0.999626 \\ 
 P\left( {{Y \mathord{\left/
 {\vphantom {Y {Y_0 }}} \right.
 \kern-\nulldelimiterspace} {Y_0 }} > 0.008856} \right)\,\, = 0.999165 \\ 
 P\left( {{Z \mathord{\left/
 {\vphantom {Z {Z_0 }}} \right.
 \kern-\nulldelimiterspace} {Z_0 }} > 0.008856} \right)\,\, = 0.997184 \\ 
 \end{array}
\end{equation}

\begin{figure}[!ht]
\centering
 \includegraphics[width=0.36\columnwidth,draft=false]{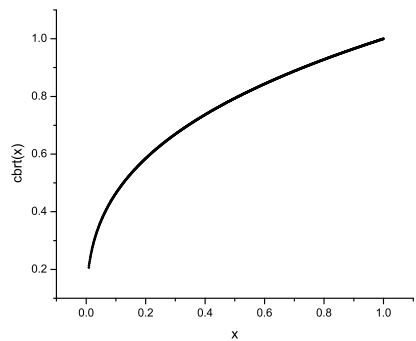}
 \caption{Cube-root function in the interval $[0.008856,1]$}
 \label{fig_cbrt}
\end{figure}

These high values suggest that we can accelerate the transformation substantially if we can devise a fast approximation for the cube-root function (see Figure \ref{fig_cbrt}). Table \ref{tab_cbrt_poly} shows the coefficients of the minimax polynomials of various degrees. Here, $n$ and $\varepsilon^A_{max}$ represent the degree of the polynomial and error of the minimax approximation, respectively.

\begin{table}
\centering
\scriptsize
{
\caption{ \label{tab_cbrt_poly} Minimax polynomials for the cube-root function }
\begin{tabular}{ c|c|c|c|c|c|c|c }
\hline
$n$ & $\varepsilon^A_{max}$ & $a_0$ & $a_1$ & $a_2$ & $a_3$ & $a_4$ & $a_5$\\
\hline
\hline
2 & 1.271154e-01 & 1.268979e-01 & 2.393873 & -1.647669 & & &\\
\hline
3 & 9.787829e-02 & 9.787826e-02 & 4.057495 & -7.388864 & 4.331370 & &\\
\hline
4 & 8.111150e-02 & 8.111133e-02 & 5.926004 & -2.017165e+01 & 2.833070e+01 & -1.324728e+01 &\\
\hline
5 & 7.002956e-02 & 7.002910e-02 & 7.961214 & -4.329352e+01 & 1.063182e+02 & -1.135685e+02 & 4.358268e+01\\
\hline
\end{tabular}
}
\end{table}

It can be seen that the error values are quite high and as the approximation degree is increased, the accuracy doesn't improve significantly. This suggests that rational functions might be better suited for this approximation task. Table \ref{tab_cbrt_rat} shows the coefficients of the minimax rationals of various degrees. Here, each pair of adjacent rows corresponds to a rational function of a particular degree $(n,m)$ in which the first and second rows represent the numerator and denominator, respectively. It can be seen that minimax rationals can accurately represent the cube-root function.

\begin{table}
\centering
\scriptsize
{
\caption{ \label{tab_cbrt_rat} Minimax rationals for the cube-root function }
\begin{tabular}{ c|c|c|c|c|c|c|c }
\hline
$n$ & $m$ & $\varepsilon^A_{max}$ & $a_0$ & $a_1$ & $a_2$ & $a_3$ & $a_4$\\
\hline
\hline
\multirow{2}{*}{2} & \multirow{2}{*}{2} &  \multirow{2}{*}{2.060996e-03} & 6.309655e-03 & 5.785782e-01 & 1.591005 & &\\
& & & 4.482646e-02 & 1.175862 & 9.596879e-01 & &\\
\hline
\multirow{2}{*}{2} & \multirow{2}{*}{3} & \multirow{2}{*}{7.210231e-04} & 2.500705e-03 & 3.447113e-01 & 1.942708 & &\\
& & & 1.978701e-02 & 8.542797e-01 & 1.664540 & -2.503267e-01 &\\
\hline
\multirow{2}{*}{3} & \multirow{2}{*}{2} & \multirow{2}{*}{5.931593e-04} & 1.776519e-03 & 2.632323e-01 & 1.751297 & 3.836709e-01 &\\
& & & 1.432256e-02 & 6.779998e-01 & 1.706260 & &\\
\hline
\multirow{2}{*}{2} & \multirow{2}{*}{4} & \multirow{2}{*}{3.107735e-04} & 1.317899e-03 & 2.390113e-01 & 2.099395 & &\\
& & & 1.124254e-02 & 6.726679e-01 & 2.184656 & -7.511150e-01 & 2.229717e-01\\
\hline
\multirow{2}{*}{3} & \multirow{2}{*}{3} & \multirow{2}{*}{1.858694e-04} & 4.370889e-04 & 9.526952e-02 & 1.252009 & 1.302733 &\\
& & & 3.912364e-03 & 2.954084e-01 & 1.717143 & 6.343408e-01 &\\
\hline
\multirow{2}{*}{4} & \multirow{2}{*}{2} & \multirow{2}{*}{2.334688e-04} & 7.589302e-04 & 1.519784e-01 & 1.663584 & 8.368075e-01 & -1.657269e-01\\
& & & 6.644723e-03 & 4.506424e-01 & 2.030622 & &\\
\hline
\multirow{2}{*}{3} & \multirow{2}{*}{4} & \multirow{2}{*}{8.539863e-05} & 1.683667e-04 & 4.667675e-02 & 9.106812e-01 & 1.810577 &\\
& & & 1.610864e-03 & 1.617974e-01 & 1.494070 & 1.218468 & -1.079451e-01\\
\hline
\multirow{2}{*}{4} & \multirow{2}{*}{3} & \multirow{2}{*}{8.052920e-05} & 1.349673e-04 & 3.832079e-02 & 7.870174e-01 & 1.799062 & 2.071170e-01\\
& & & 1.299250e-03 & 1.345420e-01 & 1.330358 & 1.365369 &\\
\hline
\multirow{2}{*}{4} & \multirow{2}{*}{4} & \multirow{2}{*}{3.856930e-05} & 3.927283e-05 & 1.392318e-02 & 4.114739e-01 & 1.734853 & 8.679223e-01\\
& & & 4.022100e-04 & 5.414536e-02 & 8.221526e-01 & 1.800167 & 3.513617e-01\\
\hline
\end{tabular}
}
\end{table}

\subsection{HSI Color Space}
\label{sec_hsi}

\subsubsection{Color Space Description}
\label{sec_hsi_desc}

HSI (Hue-Saturation-Intensity) is an intuitive alternative to the RGB space \cite{Plataniotis00}. It uses approximately cylindrical coordinates, and is a non-linear deformation of the RGB color cube. The hue H is a function of the angle in the polar coordinate system and describes a pure color. The saturation S is proportional to radial distance and denotes the purity of a color. Finally, the intensity I is the distance along the axis perpendicular to the polar coordinate plane and represents the brightness. 

The transformation between RGB and HSI is given by:

\begin{equation}
\label{equ_rgb_hsi}
\begin{array}{l}
 H = \arccos\left[ {\frac{{0.5\left( {R - G + R - B} \right)}}{{\sqrt {\left( {R - G} \right)^2  + \left( {R - B} \right)\left( {G - B} \right)} }}} \right] \\ 
 \mbox{if}\,\,\left( {B > G} \right) \\ 
 \quad H = 2\pi  - H \\ 
 S = 1 - {{3\,\min (R,G,B)} \mathord{\left/
 {\vphantom {{3\,\min (R,G,B)} {\left( {R + G + B} \right)}}} \right.
 \kern-\nulldelimiterspace} {\left( {R + G + B} \right)}}\\
 I = {{\left( {R + G + B} \right)} \mathord{\left/
 {\vphantom {{\left( {R + G + B} \right)} 3}} \right.
 \kern-\nulldelimiterspace} 3}
\end{array}
\end{equation}

where \emph{arccos} denotes the inverse cosine function. Kender \cite{Kender76} proposed a fast version of (\ref{equ_rgb_hsi}) that gives numerically identical results. This transformation involves fewer multiplications and no square root operation:

\begin{equation}
\label{equ_rgb_kender}
\begin{array}{l}
 \mbox{if}\,\,R > B\,\,\mbox{and}\,\,G > B \\ 
 \quad H = \frac{\pi }{3} + \arctan\left[ {\frac{{\sqrt 3 \left( {G - R} \right)}}{{G - B + R - B}}} \right] \\ 
 \mbox{else}\,\,\mbox{if}\,\,G > R \\ 
 \quad H = \pi  + \arctan\left[ {\frac{{\sqrt 3 \left( {B - G} \right)}}{{B - R + G - R}}} \right] \\ 
 \mbox{else}\,\,\mbox{if}\,\,B > G \\ 
 \quad H = \frac{{5\pi }}{3} + \arctan\left[ {\frac{{\sqrt 3 \left( {R - B} \right)}}{{R - G + B - G}}} \right] \\ 
 \mbox{else}\,\,\mbox{if}\,\,R > B \\ 
 \quad H = 0 \\ 
 \mbox{else} \\
 \quad H = \mbox{undefined} \\
 \end{array}
\end{equation}

where \emph{arctan} denotes the inverse tangent function. The distance between two pixels $\mathbf{x}$ and $\mathbf{y}$ in the HSI space is given by:

\begin{equation}
\label{equ_hsi_dist}
\begin{array}{l}
 D_{HSI}(\mathbf{x},\mathbf{y}) = \sqrt {s_\mathbf{x}^2  + s_\mathbf{y}^2 - 2s_\mathbf{x} s_\mathbf{y} \cos \theta + (i_\mathbf{x} - i_\mathbf{y})^2}  \\ 
 \theta  = \left\{ \begin{array}{l}
 \left| {h_\mathbf{x} - h_\mathbf{y}} \right|\quad \quad \quad \,\,\mbox{if}\,\,\left| {h_\mathbf{x} - h_\mathbf{y}} \right|\, \le \,\pi  \\ 
 2\pi  - \left| {h_\mathbf{x} - h_\mathbf{y}} \right|\quad \mbox{otherwise} \\ 
 \end{array} \right. \\ 
 \end{array}
\end{equation}

\subsubsection{Approximation of the Inverse Tangent Function}
\label{sec_hsi_arctan}

We decided to approximate Kender's transformation (\ref{equ_rgb_kender}) rather than the original one (\ref{equ_rgb_hsi}) due to several reasons. First, these two give identical results. Second, (\ref{equ_rgb_kender}) is computationally cheaper than (\ref{equ_rgb_hsi}). Third, as will be seen in the next subsection, the inverse tangent function in (\ref{equ_rgb_kender}) is easier to approximate when compared to the inverse cosine function in (\ref{equ_rgb_hsi}).

In (\ref{equ_rgb_kender}), all of the cases involve multiplication by the constant $\sqrt{3}$. Therefore, a multiplication operation can be avoided by approximating $\arctan\left(\sqrt{3}x\right)$. Note that the argument of this function can also be negative. This can be handled using the following identity:

\begin{equation}
\arctan\left( x \right) =  - \arctan\left( { - x} \right)
\end{equation}

\begin{figure}[!ht]
\centering
 \includegraphics[width=0.36\columnwidth,draft=false]{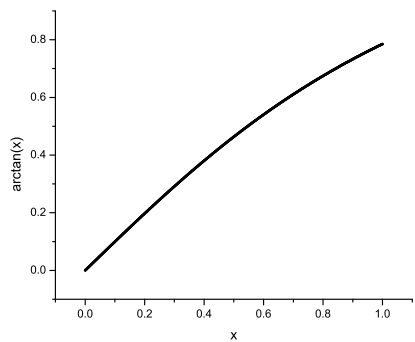}
 \caption{Inverse tangent function in the interval $[0,1]$}
 \label{fig_arctan_01}
\end{figure}

In (\ref{equ_rgb_kender}), the inverse tangent function receives its arguments from the interval $[-1.0, 1.0)$\footnote{This holds prior to the multiplication with $\sqrt{3}$.}. Figure \ref{fig_arctan_01} shows a plot of the function in the second half of this interval.

Owing to its highly linear behavior, this function can be accurately approximated by low-order polynomials. Table \ref{tab_arctan} shows the coefficients of the minimax polynomials of various degrees ($n$).

\begin{table}
\centering
\scriptsize
{
\caption{ \label{tab_arctan} Minimax polynomials for the inverse tangent function }
\begin{tabular}{ c|c|c|c|c|c|c|c }
\hline
$n$ & $\varepsilon^A_{max}$ & $a_0$ & $a_1$ & $a_2$ & $a_3$ & $a_4$ & $a_5$\\
\hline
\hline
2 & 6.907910e-03 & 5.959793e-03 & 1.782975 & 7.497879e-01 & & &\\
\hline
3 & 3.654156e-03 & -3.654076e-03 & 1.884080 & -9.805583e-01 & 1.430580e-01 & &\\
\hline
4 & 1.286371e-03 & -1.286369e-03 & 1.796716 & -4.958969e-01 & -6.927404e-01 & 4.421541e-01\\
\hline
5 & 1.801311e-04 & -1.801283e-04 & 1.739333 & -2.039848e-02 & -2.065512 & 2.052837 & -6.591729e-01\\
\hline
\end{tabular}
}
\end{table}

\subsection{Spherical Coordinate Transform (SCT) Color Space}
\label{sec_sct}

\subsubsection{Color Space Description}
\label{sec_sct_desc}

The Spherical Coordinate Transform (SCT) is defined as \cite{Umbaugh89}:

\begin{equation}
\label{equ_rgb_sct}
\begin{array}{l}
 L\,\,\,\,\, = \sqrt {R^2  + G^2  + B^2 }  \\ 
 \angle A = \arccos\left( {{B \mathord{\left/
 {\vphantom {B L}} \right.
 \kern-\nulldelimiterspace} L}} \right) \\ 
 \angle B = \arccos\left( {\frac{R}{{L\sin \left( {\angle A} \right)}}} \right) \\ 
 \end{array}
\end{equation}

where $L$ represents the luminance, and angles $\angle A$ and $\angle B$ represent the chromaticity.

The expression for $\angle B$ can be simplified using trigonometric manipulations:

\begin{equation}
\label{equ_sct_b}
\angle B = \arctan\left( {{G \mathord{\left/
 {\vphantom {G R}} \right.
 \kern-\nulldelimiterspace} R}} \right)
\end{equation}

This formulation is computationally advantageous in that it avoids a multiplication and a sine operation.

Unfortunately, it is not easy to define a perceptual distance function in SCT. Although various formulae have been developed to calculate the distance between two points lying on the same spherical surface, these cannot be used in SCT. This is because pixels of different brightness in this space lie on different spherical shells. Therefore, we decided to use the following alternative approach. Given two pixels in SCT, the inverse transformation (\ref{equ_sct_rgb}) is applied to switch back to the RGB space. The pixels are then converted to CIELAB so that the distance between them can be calculated using (\ref{equ_cielab_dist}). This indirect method of distance calculation is likely to introduce additional errors. However, minimax approximations for the elementary functions involved in (\ref{equ_rgb_sct}) and (\ref{equ_sct_b}) can be devised to obtain an arbitrarily accurate approximate transformation.

\begin{equation}
\label{equ_sct_rgb}
\begin{array}{l}
 R = L\sin \left( {\angle A} \right)\cos \left( {\angle B} \right) \\ 
 G = L\sin \left( {\angle A} \right)\sin \left( {\angle B} \right) \\ 
 B = L\cos \left( {\angle A} \right) \\ 
 \end{array}
\end{equation}

\subsubsection{Approximation of the Inverse Cosine Function}
\label{sec_sct_arccos}

\begin{figure}[!ht]
\centering
 \includegraphics[width=0.36\columnwidth,draft=false]{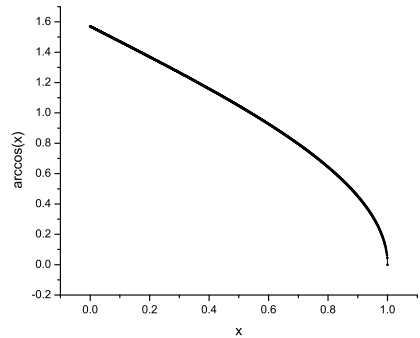}
 \caption{Inverse cosine function in the interval $[0,1]$}
 \label{fig_arccos}
\end{figure}

In the expression for $\angle A$, the inverse cosine function receives its arguments from the interval $[0,1]$. Figure \ref{fig_arccos} shows a plot of this function.

Unfortunately, approximating the inverse cosine function in this interval is not easy because of its behavior near 1. This can be circumvented using the following numerically more stable identity for $x \geq 0.5$:

\begin{equation}
\label{equ_arccos_stable}
\arccos(x) = 2 \arcsin\left( {\sqrt {0.5(1 - x)} } \right)
\end{equation}

\begin{figure}[!ht]
\centering
 \includegraphics[width=0.36\columnwidth,draft=false]{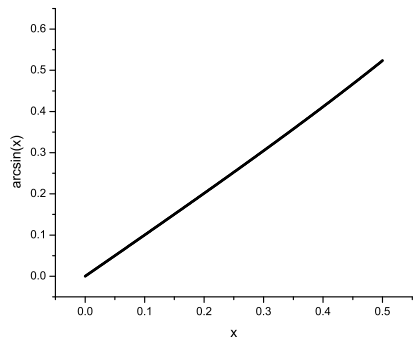}
 \caption{Inverse sine function in the interval $[0,0.5]$}
 \label{fig_arcsin}
\end{figure}

In (\ref{equ_arccos_stable}), the inverse sine function (\emph{arcsin}) receives its arguments from the interval $[0,0.5]$. Figure \ref{fig_arcsin} shows a plot of this function. In order to avoid two multiplication operations, the following function can be approximated instead:

\begin{equation}
\label{equ_arccos_cheap}
\begin{array}{l}
 y = \sqrt {1 - x} \\ 
 \arccos(x) = 2\arcsin\left( {{y \mathord{\left/
 {\vphantom {y {\sqrt 2 }}} \right.
 \kern-\nulldelimiterspace} {\sqrt 2 }}} \right) \\ 
 \end{array}
\end{equation}

In (\ref{equ_arccos_cheap}), the argument $y$ falls into the interval $\,\left[ {0,\,{1 \mathord{\left/ {\vphantom {1 {\sqrt 2 }}} \right. \kern-\nulldelimiterspace} {\sqrt 2 }}} \right]$. Table \ref{tab_arcsin} shows the coefficients of the minimax polynomials of various degrees ($n$).

\begin{table}
\centering
\scriptsize
{
\caption{ \label{tab_arcsin} Minimax polynomials for the inverse sine function }
\begin{tabular}{ c|c|c|c|c|c|c|c }
\hline
\hline
$n$ & $\varepsilon^A_{max}$ & $a_0$ & $a_1$ & $a_2$ & $a_3$ & $a_4$ & $a_5$\\
\hline
4 & 2.097814e-05 & 2.097797e-05 & 1.412840 & 1.429881e-02 & 6.704361e-02 & 6.909677e-02\\
\hline
5 & 2.370540e-06 & -2.370048e-06 & 1.414434 & -3.300037e-03 & 1.354670e-01 & -3.994259e-02 & 6.099502e-02\\
\hline
\end{tabular}
}
\end{table}

On the other hand, it can be seen from Figure \ref{fig_arccos} that the inverse cosine function is highly linear in the interval $[0,0.5]$ and can be accurately approximated by polynomials. Table \ref{tab_arccos} shows the coefficients of the minimax polynomials of various degrees ($m$).

\begin{table}
\centering
\scriptsize
{
\caption{ \label{tab_arccos} Minimax polynomials for the inverse cosine function }
\begin{tabular}{ c|c|c|c|c|c|c|c }
\hline
$m$ & $\varepsilon^A_{max}$ & $a_0$ & $a_1$ & $a_2$ & $a_3$ & $a_4$ & $a_5$\\
\hline
\hline
4 & 1.048949e-05 & 1.570786 & -9.990285e-01 & -1.429899e-02 & -9.481335e-02 & -1.381942e-01\\
\hline
5 & 1.186403e-06 & 1.570798 & -1.000156 & 3.299810e-03 & -1.915780e-01 & 7.988231e-02 & -1.725177e-01\\
\hline
\end{tabular}
}
\end{table}

\subsubsection{Approximation of the Inverse Tangent Function}
\label{sec_sct_arctan}

\begin{figure}[!ht]
\centering
 \includegraphics[width=0.36\columnwidth,draft=false]{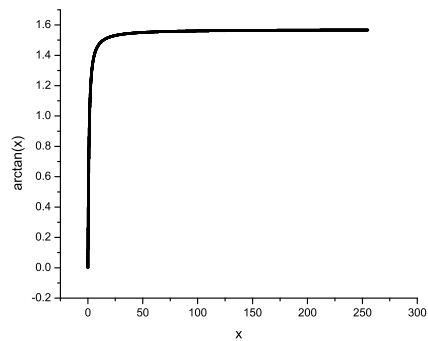}
 \caption{Inverse tangent function in the interval $[0,255]$}
 \label{fig_arctan0255}
\end{figure}

In (\ref{equ_sct_b}), the inverse tangent function receives its arguments from the interval $[0,255]$. Figure \ref{fig_arctan0255} shows a plot of this function. Note that the degenerate cases, i.e.\ $R = G = 0 \,\, (\angle B = \mbox{undefined})$ and $R = 0,\, G > 0 \,\, (\angle B = \pi / 2 )$, are not reflected in the plot.

It can be seen that the function exhibits high variability in this large interval. In order to obtain an accurate low-order polynomial approximation, the domain can be divided into two as follows:

\begin{equation}
\label{arctan_split}
\begin{array}{l}
 x = {G \mathord{\left/
 {\vphantom {G R}} \right.
 \kern-\nulldelimiterspace} R} \\ 
 \arctan(x) = \left\{ \begin{array}{l}
 f(x) = \arctan(x)\quad \quad \,\,\quad \,\,\,\,\,\,\, \mbox{if} \,\, G < R \\ 
 g(x) = {\pi  \mathord{\left/
 {\vphantom {\pi  2}} \right.
 \kern-\nulldelimiterspace} 2} - \arctan\left( {{1 \mathord{\left/
 {\vphantom {1 x}} \right.
 \kern-\nulldelimiterspace} x}} \right)\,\,\,\mbox{otherwise} \\ 
 \end{array} \right. \\ 
 \end{array}
\end{equation}

The second part of (\ref{arctan_split}) follows from:

\begin{equation}
\arctan(x) = {\pi  \mathord{\left/
 {\vphantom {\pi  2}} \right.
 \kern-\nulldelimiterspace} 2} - \arctan\left( {{1 \mathord{\left/
 {\vphantom {1 x}} \right.
 \kern-\nulldelimiterspace} x}} \right)\quad \mbox{for}\,\,x > 0
\end{equation}

When $G < R$, the function $f$ receives its arguments from the interval $\,\left[ {0,{{254} \mathord{\left/ {\vphantom {{254} {255}}} \right. \kern-\nulldelimiterspace} {255}}} \right]$. On the other hand, when $R \leq G$ the inverse tangent function within $g$ receives its arguments from the interval  $\,\left[ {1/255,1} \right]$. Tables \ref{tab_f} and \ref{tab_g} show the coefficients of the minimax polynomials of various degrees ($r$) for the functions $f$ and $g$, respectively.

\begin{table}
\centering
\scriptsize
{
\caption{ \label{tab_f} Minimax polynomials for the function $f$ }
\begin{tabular}{ c|c|c|c|c|c|c|c }
\hline
$r$ & $\varepsilon^A_{max}$ & $a_0$ & $a_1$ & $a_2$ & $a_3$ & $a_4$ & $a_5$\\
\hline
\hline
4 & 1.036515e-04 & -1.036508e-04 & 1.003740 & -1.773538e-02 & -3.390563e-01 & 1.386796e-01\\
\hline
5 & 2.073939e-05 & 2.073866e-05 & 9.982666e-01 & 2.352573e-02 & -4.506862e-01 & 2.635050e-01 & -4.920822e-02\\
\hline
\end{tabular}
}
\end{table}

\begin{table}
\centering
\scriptsize
{
\caption{ \label{tab_g} Minimax polynomials for the function $g$ }
\begin{tabular}{ c|c|c|c|c|c|c|c }
\hline
$r$ & $\varepsilon^A_{max}$ & $a_0$ & $a_1$ & $a_2$ & $a_3$ & $a_4$ & $a_5$\\
\hline
\hline
4 & 1.051643e-04 & 1.570917 & -1.004004 & 1.885694e-02 & 3.373159e-01 & -1.377930e-01\\
\hline
5 & 2.012104e-05 & 1.570769 & -9.981253e-01 & -2.440212e-02 & 4.528921e-01 & -2.659181e-01 & 5.016228e-02\\
\hline
\end{tabular}
}
\end{table}

As in the case of (\ref{equ_cbrt_prob}), the probabilities of calling the inverse sine and inverse cosine functions can be calculated from an image that contains every possible color in the 24-bit RGB space:

\begin{equation}
\label{equ_sct_arcsin_arccos_prob}
\begin{array}{l}
 P_{\arcsin}  = P\left( {{B \mathord{\left/
 {\vphantom {B L}} \right.
 \kern-\nulldelimiterspace} L} \ge 0.5} \right) = 0.557723 \\ 
 P_{\arccos} = 1 - P_{{\rm arcsin}} = 0.442277 \\ 
 \end{array}
\end{equation}

In contrast, the inverse tangent function in (\ref{equ_sct_b}) is almost always called (except for when $R = G = 0$, in which case  $\angle B = \mbox{undefined}$):

\begin{equation}
\label{equ_sct_b_prob}
P_{\arctan}  = 1 - P\left( {R = G = 0} \right)\, = 1 - \frac{{256}}{{256^3 }} = 0.999985
\end{equation}

Note that $\angle B = \pi/2$ when $R = 0$ and $G > 0$.

\subsection{Experimental Results}
\label{sec_exp}

\subsubsection{Calculation of the Color Space Transformation Errors}
\label{sec_exp_err}

In order to calculate the accuracy of the presented approximate color space transformations, we used a $4,096 \times 4,096$ image (henceforth referred to as \emph{RGB16Million})\footnote{Available at \url{http://brucelindbloom.com/downloads/RGB16Million.tif.zip}}, which contains $16,777,216$ unique colors, i.e.\ every possible color in the 24-bit RGB space. Tables \ref{tab_cielab}-\ref{tab_sct} show the average execution time (in seconds)\footnote{Programming language: C, Compiler: gcc 3.4.4, CPU: Intel Pentium D 2.66Ghz} over $1,000$ identical runs and the average ($\varepsilon^T_{avg}$) and maximum ($\varepsilon^T_{max}$) transformation errors for CIELAB (\ref{equ_cielab_dist}), HSI (\ref{equ_hsi_dist}), and SCT (\ref{equ_cielab_dist}) \& (\ref{equ_sct_rgb}), respectively. Note that the rows of Table \ref{tab_cielab} are sorted on ($n + m$), since the execution time is proportional to the total degree of the polynomials in the rational. On the other hand, the rows of Table \ref{tab_sct} are sorted on the degree of the inverse tangent approximation ($r$). This is because the probability of calling this function (\ref{equ_sct_b_prob}) is much higher than that of the other two elementary functions (\ref{equ_sct_arcsin_arccos_prob}). Therefore, the execution time is mainly influenced by the degree of the inverse tangent approximation.

We also performed the exact transformations on \emph{RGB16Million} and calculated the average execution time over 100 runs. The results were $60.331$s, $13.485$s, and $27.599$s for CIELAB, HSI, and SCT, respectively. Comparing these values with those given in Tables \ref{tab_cielab}-\ref{tab_sct}, we can see that the proposed approximations provide substantial computational savings.

\subsubsection{Calculation of the Computational Gain Values}
\label{sec_exp_gain}

It can be seen from Tables \ref{tab_cielab}-\ref{tab_sct} that only marginal computational gains can be obtained using lower-order approximations without significantly compromising the accuracy of the transformation. Therefore, in this subsection, we consider only the  highest-order approximations for each transformation. However, lower-order approximations might be preferable depending on the application requirements.

In order to calculate the computational gain for each approximate transformation, a set of $100$ high quality RGB images was collected from the Internet. The set includes images of people, animals, plants, buildings, aerial maps, man-made objects, natural scenery, paintings, sketches, as well as scientific, biomedical, synthetic images and test images commonly used in the literature. 

On each image, the exact transformations were performed $100$ times and the average execution times were calculated. The same was done for the corresponding approximate transformations with $1,000$ runs. The computational gain is calculated as the ratio of the average execution times. Table \ref{tab_gain} shows the statistics over the entire image set. It can be seen that the computational gain observed in \emph{RGB16Million} also applies to the case of real-world images.

\begin{table}
\centering
\caption{ \label{tab_cielab} Comparison of CIELAB approximations }
\begin{tabular}{ c|c|c|c|c }
\hline
$n$ & $m$ & Avg.\ Time & $\varepsilon^T_{avg}$ & $\varepsilon^T_{max}$\\
\hline
\hline
2 & 2 & 1.031872 & 0.787741 & 2.225399\\
\hline
2 & 3 & 1.134568 & 0.295577 & 0.774404\\
\hline
3 & 2 & 1.114776 & 0.239033 & 0.636929\\
\hline
2 & 4 & 1.221390 & 0.131411 & 0.331015\\
\hline
3 & 3 & 1.165900 & 0.061127 & 0.193565\\
\hline
4 & 2 & 1.197828 & 0.090846 & 0.247291\\
\hline
3 & 4 & 1.291526 & 0.018092 & 0.085046\\
\hline
4 & 3 & 1.291570 & 0.014927 & 0.079935\\
\hline
4 & 4 & 1.292872 & 0.003201 & 0.036481\\
\hline
\end{tabular}
\end{table}

\begin{table}
\centering
\caption{ \label{tab_hsi} Comparison of HSI approximations }
\begin{tabular}{ c|c|c|c }
\hline
$n$ & Avg.\ Time & $\varepsilon^T_{avg}$ & $\varepsilon^T_{max}$\\
\hline
\hline
2 & 0.865328 & 0.002523 & 0.007134\\
\hline
3 & 0.865575 & 0.001259 & 0.003693\\
\hline
4 & 0.866254 & 0.000448 & 0.001320\\
\hline
5 & 0.873251 & 0.000063 & 0.000190\\
\hline
\end{tabular}
\end{table}

\begin{table}
\centering
\caption{ \label{tab_sct} Comparison of SCT approximations }
\begin{tabular}{ c|c|c|c|c|c }
\hline
$n$ & $m$ & $r$ & Avg.\ Time & $\varepsilon^T_{avg}$ & $\varepsilon^T_{max}$\\
\hline
\hline
4 & 4 & 4 & 1.330927 & 0.006596 & 0.024273\\
\hline
4 & 5 & 4 & 1.371811 & 0.006553 & 0.024273\\
\hline
5 & 4 & 4 & 1.377924 & 0.006266 & 0.022820\\
\hline
5 & 5 & 4 & 1.419315 & 0.006224 & 0.021810\\
\hline
4 & 4 & 5 & 1.427337 & 0.002139 & 0.008371\\
\hline
4 & 5 & 5 & 1.462627 & 0.002011 & 0.008371\\
\hline
5 & 4 & 5 & 1.471486 & 0.001381 & 0.005793\\
\hline
5 & 5 & 5 & 1.507812 & 0.001254 & 0.004543\\
\hline
\end{tabular}
\end{table}

\begin{table}
\centering
\caption{ \label{tab_gain} Computational gain statistics for each transformation }
\begin{tabular}{ c|c|c|c|c|c }
\hline
\multirow{2}{*}{Transformation} & \multicolumn{5}{|c}{Computational Gain}\\
& Min & Max & Mean & Stdev & Median\\
\hline
\hline
CIELAB & 8.761736 & 40.629135 & 35.575373 & 4.608590 & 37.192729\\
\hline
HSI & 4.511637 & 16.460248 & 11.437106 & 2.439482 & 11.669715\\
\hline
SCT & 4.493543 & 22.232897 & 15.782453 & 2.110820 & 16.162434\\
\hline
\end{tabular}
\end{table}

\subsubsection{Comparison with 3D Lookup Table Interpolation Methods}
\label{sec_exp_lut}

In this subsection we compare MACT with the most commonly used LUT interpolation methods, namely trilinear, prism, pyramidal, and tetrahedral. These methods involve three steps: packing, extraction, and interpolation \cite{Kang06}. Packing is a process that divides the domain of the source space and populates it with sample points to build the LUT. The extraction step aims at finding the location of the input pixel and extracting the color values of the nearest lattice points. The last step is 3D interpolation, in which the input point and the extracted lattice points are used to calculate the destination color specifications. Essentially, 3D interpolation is a repeated application of linear interpolation \cite{Kang06}:
 
\begin{equation}
x = (1 - \alpha )x_0  + \alpha x_1 
\end{equation}

where $x_0$ and $x_1$ are the spatial coordinates of the two known points and $\alpha$ is the interpolation coefficient. The 3D interpolation is the step in which the aforementioned LUT interpolation methods differ. Trilinear interpolation uses $8$ neighboring lattices, whereas prism, pyramidal, and tetrahedral interpolations use $6$, $5$, and $4$ neighbors, respectively. The greater the number of neighboring lattices used in a method, the higher the computational requirements and accuracy. Since the formulations of these methods are mathematically involved, the interested reader is referred to the relevant literature \cite{Kang06}.

We implemented two versions of each LUT method: standard and caching. The former is a direct implementation of the mathematical formulation, whereas the latter pre-computes the differences between neighboring lattices and stores these in each node. Table \ref{tab_lut} shows the average execution times on \emph{RGB16Million} over $1,000$ runs and the transformation errors for CIELAB. It can be seen that even though these methods range from being $1.32$ to $2.87$ times faster than MACT, their accuracy is $18$ to $161$ times lower. In addition, these methods require extra storage of up to $10.7$ MBs. Note that LUT size only affects the accuracy of the transformation and the storage requirements and, in theory, it should not affect the computational time. However, it can be observed from the table that as the LUT size is increased, the computational time increases as well. This is most likely due to the limited size of the processor cache.

\begin{table}
\centering
\caption{ \label{tab_lut} Comparison of 3D LUT interpolation methods }
\begin{tabular}{ c|c|c|c|c|c }
\hline
Method & LUT & Avg.\ Time & Avg.\ Time & $\varepsilon^T_{avg}$ & $\varepsilon^T_{max}$\\
& Size & (standard) & (caching) & &\\
\hline
\hline
\multirow{3}{*}{Trilinear} & $9^3$ & 0.901673 & 0.619681 & 0.359677 & 5.409771\\
 & $17^3$ & 0.908753 & 0.630465 & 0.100026 & 1.987040\\
 & $33^3$ & 0.982880 & 0.716958 & 0.025089 & 0.653506\\
\hline
\multirow{3}{*}{Prism} & $9^3$ & 0.856652 & 0.609452 & 0.349128 & 5.409771\\
 & $17^3$ & 0.860880 & 0.614681 & 0.095150 & 2.343905\\
 & $33^3$ & 0.931349 & 0.677319 & 0.024172 & 0.826651\\
\hline
\multirow{3}{*}{Pyramidal} & $9^3$ & 0.850409 & 0.540481 & 0.319163 & 5.872591\\
 & $17^3$ & 0.872813 & 0.566271 & 0.088116 & 2.909481\\
 & $33^3$ & 0.940143 & 0.637049 & 0.022466 & 1.042665\\
\hline
\multirow{3}{*}{Tetrahedral} & $9^3$ & 0.689341 & 0.450413 & 0.284339 & 5.783921\\
 & $17^3$ & 0.704121 & 0.486654 & 0.077870 & 2.702448\\
 & $33^3$ & 0.773102 & 0.561076 & 0.020253 & 1.185788\\
\hline
\end{tabular}
\end{table}

As for HSI and SCT, since they include angular components (H component in HSI, $\angle A$ and $\angle B$ components in SCT), 3D interpolation in these color spaces involves the interpolation of angular (circular) data given by \cite{Mardia01}:

\begin{equation}
\label{equ_angular}
\theta  = \arctan\left( {\frac{{(1 - \alpha )\,{\rm sin}\left( {\theta _0 } \right) + \alpha \,{\rm sin}\left( {\theta _1 } \right)}}{{(1 - \alpha )\,{\rm cos}\left( {\theta _0 } \right) + \alpha \,\cos \left( {\theta _1 } \right)}}} \right)\,
\end{equation}

where $\theta_0$ and $\theta_1$ are the two known angles and $\alpha$ is the interpolation coefficient. Due to the trigonometric and inverse-trigonometric functions in (\ref{equ_angular}), it turns out that 3D interpolation in these spaces is computationally more expensive than the original transformations given in (\ref{equ_rgb_hsi}) and (\ref{equ_rgb_sct}). For example, trilinear interpolation in HSI takes about $84$s on \emph{RGB16Million} ($\varepsilon^T_{avg} = 0.000777$, $\varepsilon^T_{max} = 1.086255$).

Note that even if the angular data was quantized into, say $360$ steps, precomputing (\ref{equ_angular}) would not be possible since the value of the interpolation coefficient $\alpha$ is not known \emph{a priori}. In addition, the inverse tangent function cannot be approximated using a minimax polynomial since its argument is not bounded. In short, the expensive interpolation operation offsets the computational advantage of 3D LUT interpolation in angular color spaces.

\section{Conclusions}
\label{sec_conc}

In this paper we proposed MACT, a novel approach to speed up color space transformations based on minimax approximations. Advantages of MACT include ease of implementation, negligible memory requirements, extremely good accuracy, and very high computational speed. Comparisons with commonly used 3D LUT interpolation methods revealed that MACT yields significantly more accurate transformations at the expense of slightly higher computational requirements. Although MACT was applied to three particular color space transformations, it can easily be adapted to other transformations, e.g.\ RGB to CIELUV transformation, that involve computationally expensive mathematical functions.

Implementations of the fast color space transformations described in this paper will be made publicly available at \url{http://sourceforge.net/projects/fourier-ipal}.

\section*{Acknowledgments}
This publication was made possible by a grant from The Louisiana Board of Regents (LEQSF2008-11-RD-A-12). The authors are grateful to the anonymous reviewers for their valuable comments and to Bruce Lindbloom for providing the \emph{RGB16Million} image.

\bibliographystyle{IEEEbib}
\bibliography{color_spaces_bib}

\end{document}